\documentclass[letterpaper, 10 pt, conference]{ieeeconf}  

\IEEEoverridecommandlockouts                              
\let\labelindent\relax

\usepackage{times}
\usepackage{amsmath,amsfonts}
\usepackage{algorithmic}
\usepackage{algorithm}
\usepackage{array}
\usepackage{balance}
\usepackage[dvipsnames,table]{xcolor}
\usepackage{textcomp}
\usepackage{stfloats}
\usepackage{url}
\usepackage{verbatim}
\usepackage{graphicx}
\usepackage{todonotes}
\usepackage{booktabs} 
\usepackage{bbm} 
\usepackage{mathtools}
\usepackage{pgf} 
\usepackage{etoolbox} 
\usepackage{enumitem} 
\usepackage{amssymb}

\usepackage{multicol}
\usepackage{hyperref} 
\usepackage{lipsum} 
\usepackage{siunitx} 
\usepackage{mathtools}
\usepackage{xspace}
\usepackage{annotate-equations} 

\usepackage{multirow}
\usepackage{pbox}

\usepackage{threeparttable}
\usepackage{tablefootnote}

\usepackage{tikz}

\usepackage[normalem]{ulem}
\usepackage{gensymb}

\definecolor{MK_Two_One}{RGB}{178,24,43} 
\definecolor{MK_Two_Two}{RGB}{239,138,98}
\definecolor{MK_Two_Three}{RGB}{253,219,199}
\definecolor{MK_Two_Four}{RGB}{209,229,240}
\definecolor{MK_Two_Five}{RGB}{103,169,207}
\definecolor{MK_Two_Six}{RGB}{33,102,172} 

\hypersetup{
colorlinks=true
,linkcolor=black
,citecolor=black
,filecolor=MK_Two_Six
,urlcolor= MK_Two_Six
,menucolor=MK_Two_Five
,runcolor=MK_Two_Four
,linkbordercolor=MK_Two_One
,citebordercolor=MK_Two_Two
,filebordercolor=MK_Two_Three
,urlbordercolor=MK_Two_Six
,menubordercolor=MK_Two_Five
,runbordercolor=MK_Two_Four
}

\usepackage{subcaption}
\definecolor{high}{RGB}{116, 173, 209}  
\definecolor{low}{RGB}{244, 109, 67}  

\def\figref#1{Fig.~\ref{#1}}
\def\tabref#1{Tab.~\ref{#1}}
\def\eqref#1{Eq.~(\ref{#1})}

\usepackage{tikz}

\usepackage[abbreviations]{glossaries-extra}

\glssetcategoryattribute{abbreviation}{indexonlyfirst}{true}

\glssetcategoryattribute{abbreviation}{nohyperfirst}{true}

\newabbreviation{auroc}{AUROC}{Area Under the Receiver Operating Characteristic Curve}
\newabbreviation{accuracy}{Acc}{Accuracy}
\newabbreviation{ate}{ATE}{Absolute Trajectory Error}

\newabbreviation{clip}{CLIP}{Contrastive Language-Image Pretraining}
\newabbreviation{cnn}{CNN}{Convolutional Neural Network}

\newabbreviation{dof}{DoF}{DoF}

\newabbreviation{fov}{FoV}{Field of View}
\newabbreviation{fpr}{FPR}{False Positive Ratio}

\newabbreviation{gnn}{GNN}{Graph Neural Network}
\newabbreviation{gcn}{GCN}{Graph Convolutional Network}

\newabbreviation{knn}{KNN}{K-Nearest Neighbors}

\newabbreviation[plural=LLMs,firstplural=Large Language Models]{llm}{LLM}{Large Language Model}
\newabbreviation{lidar}{LiDAR}{Light Detection and Ranging}

\newabbreviation{mlp}{MLP}{Multi-Layer Perceptron}
\newabbreviation{mpc}{MPC}{Model Predictive Controller}
\newabbreviation{mse}{MSE}{Mean Squared Error}

\newabbreviation{ood}{OOD}{out-of-distribution}

\newabbreviation{rbf}{RBF}{Radial Basis Function}
\newabbreviation{rmp}{RMP}{Riemannian Motion Policies}
\newabbreviation{ros}{ROS}{Robot Operating System}
\newabbreviation{ros1}{ROS~1}{Robot Operating System}
\newabbreviation{roc}{ROC}{Receiver Operating Characteristic}
\newabbreviation{rf}{RF}{Random Forest}

\newabbreviation{sdf}{SDF}{Signed Distance Field}
\newabbreviation{slam}{SLAM}{Simultaneous Localization and Mapping}
\newabbreviation{svm}{SVM}{Support Vector Machine}
\newabbreviation{svc}{SVC}{Support Vector Classifier}
\newabbreviation{wvn}{WVN}{Wild Visual Navigation}

\newabbreviation{vit}{ViT}{Vision Transformer}
\newabbreviation{vpr}{VPR}{Visual Place Recognition}

\newcommand{\pose}[3]{\mathbf{T}_{\mathtt{#1 #2}}_{#3}}
\newcommand{\rot}[3]{\mathbf{R}_{\mathtt{#1 #2}}_{#3}}
\newcommand{\pos}[2]{{\mathtt{_#1}} \mathbf{p}_{#2}}

\newcommand{\K}{\mathbf{K}_{3\times3}}

\newcommand{\img}[1]{\mathbf{I}^{#1}}

\newcommand{\feat}[1]{\mathbf{F}^{#1}}

\newcommand{\loss}[1]{\mathcal{L}_{\mathrm{#1}}}

\newcommand{\fun}[2]{f_{\mathrm{#1}}\left( #2 \right) }

\robustify{\pose}
\robustify{\rot}
\robustify{\pos}
\robustify{\K}
\robustify{\loss}
\robustify{\feat}
\robustify{\img}
\robustify{\fun}

\definecolor{TraversableBlue}{RGB}{49, 54, 149}
\definecolor{UntraversableRed}{RGB}{192, 26, 38}
\definecolor{PaperOrange}{RGB}{251, 151, 39}
\definecolor{PaperMagenta}{RGB}{150, 36, 145}
\definecolor{PaperBlue}{RGB}{67, 110, 176}
\definecolor{PaperCyan}{RGB}{66, 173, 187}

\usepackage{amssymb}

\newcommand{\orangesquare}{{\textcolor{PaperOrange}{$\blacksquare$}}}

\usepackage{color}
\usepackage{xcolor}
\newcommand{\redsquare}{{\textcolor{red}{$\blacksquare$}}}

\begin{document}
\title{ \LARGE \bf Language-EXtended Indoor SLAM (LEXIS): \\ A Versatile System
for Real-time Visual Scene Understanding
}

\author{Christina Kassab, Matias Mattamala, Lintong Zhang, and Maurice Fallon
\thanks{The authors are with the Oxford Robotics Institute at the University of
Oxford, UK. {\tt\small \{christina, matias, lintong, mfallon\}@robots.ox.ac.uk}
} }

\hyphenation{a-ccu-ra-cy}


\maketitle

\bstctlcite{IEEEexample:BSTcontrol}

\begin{abstract}
Versatile and adaptive semantic understanding would enable autonomous systems to
comprehend and interact with their surroundings. Existing fixed-class models
limit the adaptability of indoor mobile and assistive autonomous systems. In
this work, we introduce LEXIS, a real-time indoor \gls{slam} system that
harnesses the open-vocabulary nature of \glspl{llm} to create a unified approach
to scene understanding and place recognition. The approach first builds a
topological \gls{slam} graph of the environment (using visual-inertial odometry)
and embeds \gls{clip} features in the graph nodes. We use this representation
for flexible room classification and segmentation, serving as a basis for
room-centric place recognition. This allows loop closure searches to be
directed towards semantically relevant places. Our proposed system is evaluated
using both public, simulated data and real-world data, covering office and home
environments. It successfully categorizes rooms with varying layouts and
dimensions and outperforms the state-of-the-art (SOTA). For place recognition
and trajectory estimation tasks we achieve equivalent performance to the SOTA,
all also utilizing the same pre-trained model. Lastly, we demonstrate the
system's potential for planning. Video at: \url{https://youtu.be/gRqF3euDfX8}
\end{abstract}

\section{Introduction}

Scene understanding is a long-standing problem in robot perception. Over the
last decade, \gls{slam} systems have shifted from building purely geometric
representations for localization, to semantic and interpretable representations
for interaction~\cite{Cadena2016,Davison2018}. Semantic \gls{slam} and
object-based perception have made significant advances --- powered by progress
in the machine learning and computer vision communities. \emph{3D scene
graphs}~\cite{Armeni2019,Kim2019} have more recently emerged as a
unifying representation to integrate structure and
semantics~\cite{Hughes2022,Bavle2023}. Nonetheless, the usage of fixed-class
semantic models in these applications limits the versatility of these systems.

The progress of LLM research offers a solution to this challenge, as they can
bridge the gap between visual and textual information with their open
vocabularies. Methods such as \gls{clip}~\cite{Radford2021} and
ViLD~\cite{Gu2022} have been used to enrich 3D reconstructions with semantics,
as demonstrated by methods such as OpenScene \cite{Peng2023}, ConceptFusion
\cite{Jatavallabhula2023}, and NLMap \cite{Chen2023}. These methods can identify
objects and scene properties; and can even carry out navigation using human
instructions \cite{Huang2023}. However, open questions remain about 
integrating this capability into the modules of a robotic system. In particular,
can embedded semantic understanding be harnessed for tasks such as place
recognition and localization?

\begin{figure}[t]
    \centering
    \includegraphics[width=1\columnwidth, clip, trim={0 0 0 0}]{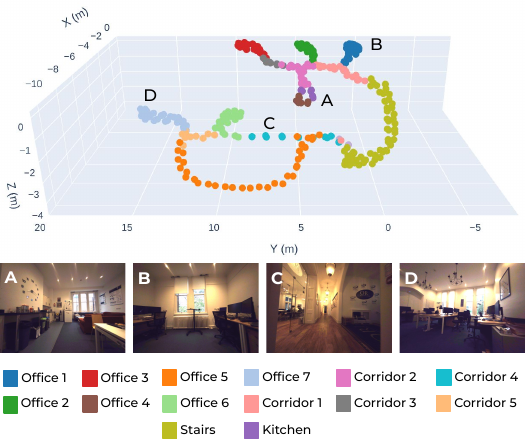}
    \caption{\small LEXIS enables pose graph segmentation from natural language.
    By exploiting the open-vocabulary capabilities of \gls{clip}, we can segment
    room instances such as \texttt{office}, \texttt{kitchen}, and
    \texttt{corridor} directly from the pose graph without fine-tuning. The
    above dataset is from a two floor office environment and contains 7 rooms as
    well as 2 corridors and stairs.}
    \label{fig:main_fig}
    \vspace{-15pt}
\end{figure}
In this work, we combine the open-vocabulary capabilities of \glspl{llm} with
classical localization and mapping methods to develop LEXIS (Language-EXtended
Indoor \gls{slam}). Unlike conventional approaches which employ separate models
for room classification, place recognition and semantic understanding, our
approach uses a single pre-trained model to efficiently execute all of these
functions. The output is a semantically segmented pose graph as shown in
\figref{fig:main_fig}. Our specific contributions are: 
\begin{itemize}
    \item A lightweight topological pose graph representation embedded with
    \gls{clip} features.
    \item A method to leverage semantic features to achieve online room
    segmentation, capable of accomodating different room sizes, layouts and
    open-floor plans. 
    \item A place recognition approach building on these room segmentations
    to propose hierarchical, room-aware loop closures. 
    \item Extensive evaluation of the system for indoor real-time room
    segmentation and classification, place recognition, and as a unified visual
    \gls{slam} system using standard and custom multi-floor datasets, with a
    demonstration for planning tasks.
\end{itemize}
\begin{figure}[t]
    \vspace{5pt}
    \centering
    \includegraphics[width=\columnwidth]{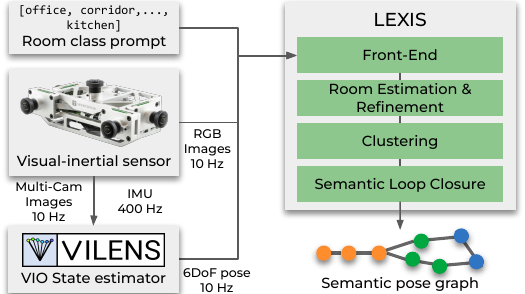}
    \caption{\small{LEXIS system overview:  The only inputs are RGB images and
an odometry estimate from a visual-inertial state estimator, as well as a prompt
list of potential room classes. The output is a semantic pose graph that encodes room
information.}}\label{fig:system-overview}
\end{figure}
\section{Related Work}
\subsection{Semantic Scene Representations}
The use of semantic information is motivated by the limitations of purely
geometric representations to encode interpretable information and to support
higher-level tasks.

Early research used boosting and Hidden Markov Models to label indoor locations
based on vision and laser range data~\cite{Rottmann2005}. Later works shifted
towards \gls{cnn}-based methods for room classification and scene understanding.
Goeddel et~al.~\cite{Goeddel2016} used \glspl{cnn} to classify LiDAR maps into
rooms, corridors and doorways. Sunderhauf et~al.~\cite{sunderhauf2015place}
overcame the closed-set limitations of \glspl{cnn} using a series of one-vs-all
classifiers to allow recognition of new semantic classes, such as allowing
generalization of door recognition to diverse settings. 

These techniques emphasize the extraction of semantic information but do not
capture contextual and higher-level understanding, such as relationships between
objects or rooms. More recent studies are directed towards the incorporation of
these semantic attributes directly into hierarchical map models, such as 3D
scene graphs~\cite{Armeni2019, Kim2019}. The multi-layered graph represents
entities such as objects, rooms, or buildings as graph nodes, while semantic
relationships are established through graph edges. Hydra~\cite{Hughes2022}
presented a five-layered scene graph with a metric-semantic 3D mesh layer,
object and agents layers, as well as, obstacle-free locations, rooms, and
buildings. Semantics are obtained through a pretrained HR-Net~\cite{Wang2019}
and \glspl{gnn} models to encode object relationships.
S-Graphs+~\cite{Bavle2023} used a similar four-layered graph and employed
geometry-based room segmentation using free-space clusters and wall planes,
without using an explicit semantic segmentation method. 

These methods rely on fixed-class models for tasks like room classification, and
face challenges in generalizing to new environments~\cite{Goeddel2016,
Hughes2023}, leading to reduced performance and difficulties with unfamiliar
room types. Approaches like Hydra need to segment the representation prior to
classification and depend on geometric data (e.g., walls and doorways). This
limits segmentation in open-floor plans or multi-functional spaces. Moreover,
current 3D scene graph representations require multiple models for semantic
segmentation, room classification, and place recognition. This requires
extensive training data and further diminishes adaptability to varied
environments.

With LEXIS we aim to address these limitations by exploiting the information
encoded in \glspl{llm}. Their open-vocabulary features enable an arbitrary
number of classes, and allow LEXIS to adapt to diverse indoor environments
without the need for pre-training or fine-tuning. Our system does not require
geometric information to perform room segmentation, enabling us to accommodate
varying room sizes and layouts, and allowing us to segment open plan spaces
effectively. Additionally, we leverage the same model for place recognition,
thereby fully capitalizing on the capabilities of \glspl{llm} throughout the
entire \gls{slam} pipeline. 

\subsection{\gls{llm}-powered Representations} Our system is inspired by other
recent works exploiting \glspl{llm} for scene representation.
 
OpenScene~\cite{Peng2023} is an offline system that enhances 3D metric
representations using \gls{clip} visual-language features. Other approaches,
such as LERF~\cite{kerr2023lerf} and
CLIP-Fields~\cite{shafiullah2022clipfields}, embed visual-language features into
neural fields to achieve 3D semantic segmentations from open-vocabulary queries.
ConceptFusion~\cite{Jatavallabhula2023} further advances these approaches by
building a representation featuring multi-modal features from vision, audio, and
language on top of a  differentiable \gls{slam}
pipeline~\cite{Jatavallabhula2020}. 

These representations have proven useful when interacting with 3D scenes,
especially for planning and navigation tasks. Natural language commands have
been employed to guide navigation tasks in indoor environments, combining
natural language plan specifications with classical state estimation and local
planning systems for navigation~\cite{shafiullah2022clipfields, Chen2023,
Huang2023}. Other 3D scene understanding tasks, such as completing partially
observed objects and localizing hidden objects have been
explored~\cite{ha2022semantic}.

LEXIS differs from the previous methods as they heavily rely on a metric
representation of the environment, necessitating the embedding and fusion of
\glspl{llm} features into a 2D or 3D map. This fusion process often needs to be
performed offline or is limited to single-room environments. 

In contrast, our system utilizes a topological representation---a pose
graph---which streamlines feature embedding while preserving the ability to use
natural language queries for segmentation in an online manner. Moreover, it
allows us to apply well-established loop closing and pose graph optimization
techniques to handle trajectory drift effectively. 

\section{Method}
A system overview of LEXIS is presented in Figure \ref{fig:system-overview}. The
main inputs are high-frequency 6~DoF odometry (for which we use our previous
work Multi-Camera VILENS~\cite{ZhangWCF22}), a stream of wide field-of-view
(FoV) RGB images, and a list of potential room classes (for example:
\texttt{office}, \texttt{kitchen}, \texttt{corridor}). The output of the system
is a \gls{clip}-enhanced semantically-segmented topological map of the
environment. The main modules of LEXIS are explained in the following sections.

\begin{figure*}[ht]
    \centering
    \includegraphics[width=\textwidth,clip, trim={0 1.5cm 0 0}]{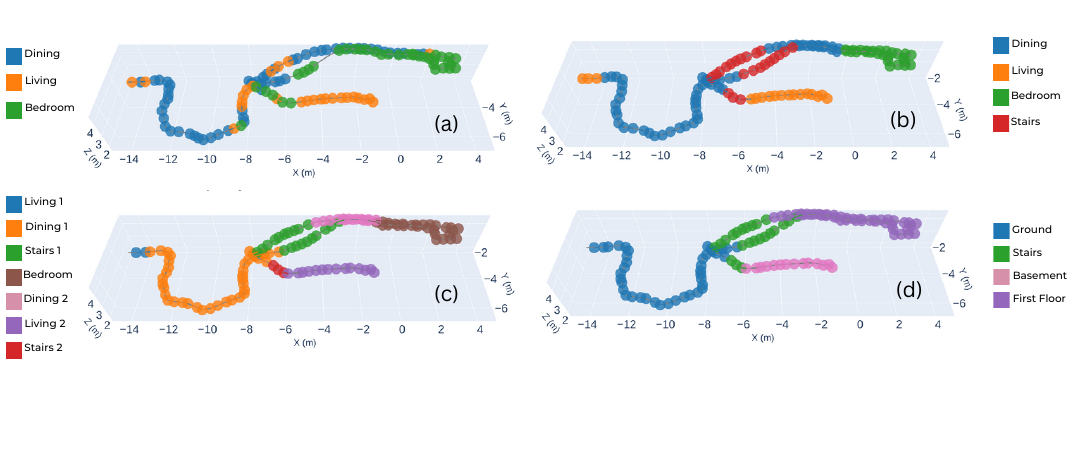}
    \caption{\small{Room segmentation and refinement on a pose graph with data
    from the uHumans2 Apartment scene (\emph{uH2-Apt}). (a) Initial room labels
    are given by CLIP. (b) The room labels post refinement. (c)
    Clustering into room instances. (d) Segmentation into
    floors.}}\label{fig:refinement}
\end{figure*}

\subsection{Front-end}
Using the high-frequency odometry estimate and RGB image stream, LEXIS builds an
incremental pose graph of equally-spaced keyframes, based on a pre-set distance
threshold. The state of the system at time $t_i$ is defined
as $\textbf{T}_{\mathtt{WB}}^i$ $\in SE(3)$, where $\mathtt{W}$ is the fixed
world frame, and $\mathtt{B}$ is the moving base frame.

As well as the pose, each node also contains \gls{clip} image encodings for
semantic understanding which we define as $\textbf{f}_{\text{CLIP}}$. We also
extract AKAZE~\cite{Alcantarilla2012} local features,
$\textbf{f}_{\text{AKAZE}}$, for loop closure registration. We extract text
encodings, denoted as $\textbf{f}_{\text{TEXT}}$, from the prior list of
potential room classes by utilizing the same \gls{clip} model. It is important
to emphasize that the room labels are not limited to predefined categories
associated with any particular dataset. 

\subsection{Room Estimation and Refinement}\label{sec:classification} As we
build the graph, we compare the image encodings, $\textbf{f}_{\text{CLIP}}$, to
the room text encodings, $\textbf{f}_{\text{TEXT}}$, using the cosine similarity
defined as:
\begin{equation}
    S_c(\mathbf{f}_{\text{CLIP}}, \mathbf{f}_{\text{TEXT}}) = \frac{\mathbf{f}_{\text{CLIP}} \cdot \mathbf{f}_{\text{TEXT}}}{{\|\mathbf{f}_{\text{CLIP}}\| \cdot \|\mathbf{f}_{\text{TEXT}}\|}} 
\end{equation}

This provides an initial room segmentation for the pose graph, as shown in
\figref{fig:refinement} (a). As this module is executed on a per-image basis
without contextual information, it can make incorrect classifications,
particularly in areas with room transitions or when images lack distinct semantic
content. To mitigate this, we employ a nearest neighbour refinement. 

The refinement approach is inspired by the Label Propagation
algorithm~\cite{Zhu2022}, a well-known technique for finding communities in
network structures. Considering the $C$ closest neighbors of each node, Label
Propagation identifies the most common label among them. If this label differs
from the node's present label, the algorithm updates the node's label. However,
in contrast to Label Propagation, which updates all the labels until
convergence, we only run one forward pass every $K$ new keyframes. In our experiments, we found that values of $C$ between 3 and 7, and $K$ between 7 and 12, produced comparable results across diverse indoor environments, including open-floor plans. 


We also use height change (in the z-axis) to detect and segment staircases, even
when they are not visibly present in the immediate surroundings. The outcome of
the full refinement module, consisting of a room label for each pose in the pose
graph, is shown in \figref{fig:refinement} (b).

\subsection{Clustering}\label{sec:clustering} Once we have allocated a room
label to each pose graph node, the next step is to group the nodes into clusters
representing individual rooms such as \texttt{office 1} and \texttt{office 2}.
For each new node with a room label not encountered previously, a new cluster is
formed. Nodes are then added to the cluster if they possess the same room label
and are within a certain distance threshold of the cluster's mean position. This
approach enables continuous updates to the clusters during the refinement
module, as allocated room labels evolve over time. 

When dealing with rooms of significantly varying sizes, it is possible for multiple clusters to emerge within a single room. In these cases, we merge clusters that have the same room label and do not have an intermediate space between them such as a corridor. The clustering outcome is presented in \figref{fig:refinement} (c). Furthermore, by identifying clusters labelled as \texttt{stairs} we can further
segment the pose graph into distinct floors (\figref{fig:refinement} (d)).


By employing this strategy, our system can organize the pose graph into a
structured representation of meaningful room instances which can also enhance
subsequent localization and loop closure modules. This adaptive clustering
approach ensures robust segmentation and can accommodate various room layouts
and sizes commonly encountered in real-world indoor environments.

\subsection{Semantic Loop Closure Detection}

The semantic information encoded in the LEXIS graph allows for efficient place
retrieval without using a dedicated place recognition model. Because of this we can reuse
this information for loop closure candidate detection.

For each new keyframe added to the graph, we first determine its corresponding
candidate room label using the image encoding, $\mathbf{f}_{\text{CLIP}}$. We then search for candidate rooms by querying all the room clusters sharing the same
label. 

We use the current localization estimate provided by the odometry to choose the
closest room cluster, and attempt geometric verification against all the
keyframes within the room using PnP \cite{Fischler1981}. For efficiency, the
query node's image encoding, $\textbf{f}_{\text{CLIP}}$, can also be compared to
nodes within the cluster using cosine similarity, further refining the candidate
set.  All successful localization attempts are then added as loop closure edges
in the pose graph, which is later optimized. The optimised poses are defined as
$\mathcal{X} := \{\mathbf{T}_{\mathtt{WB}}^1, ..., \mathbf{T}_{\mathtt{WB}}^n\}$
with the optimization formulated as a least squares minimization with a robust
DCS loss $\rho(\cdot)$ \cite{Agarwal2013}: 
\begin{equation}
    \mathcal{X} = \underset{\mathcal{X}}{\operatorname{argmin}} \sum_{i} \lVert\mathbf{r}_{\text{odom}}\rVert^2 + \sum_{i, j} \rho\left(\mathbf{r}_{\text{loop}}\right)
\end{equation}

where, $\mathbf{r}_{\text{odom}}$ refers to odometry edges and
$\mathbf{r}{_\text{loop}}$ refers to loop closures.

\section{Experiments and Results}

In this section, we demonstrate the capabilities of the system as applied to
room classification, place recognition and as a unified \gls{slam} system using
indoor real-world and simulated datasets. We conclude with a demonstration of a
mission planning application.

\subsection{Experimental Setup}
LEXIS runs in real-time on a mid-range laptop with an Intel i7 11850H @ 2.50GHz
x 16 with an Nvidia RTX A3000 GPU. The only module that requires GPU compute is
the \gls{clip} feature extractor. All other modules run on the CPU.

There are several pre-trained \gls{clip} models which use different variants of
a ResNet (RN) or a Vision Transformer (ViT) as a base. Two of the ResNet
variants follow a EfficientNet-style model scaling and use approximately 4x and
16x the compute of ResNet-50. A full list of the models evaluated are available in 
\tabref{tab:models}.

We evaluated LEXIS on three datasets:
\begin{itemize}[leftmargin=*] 
\item \emph{uHumans2} is a Unity-based simulated dataset provided by the authors
of Kimera \cite{Rosinol2021}. It has two indoor scenes: a small apartment
(\emph{uH2-Apt} [49m, 4 rooms, 3 floors]) and an office (\emph{uH2-Off} [264m, 4
rooms, 1 floor]). The dataset provides visual-inertial data, ground truth
trajectories and ground-truth bounding boxes for each room. 
\item \emph{ORI} [253m, 7 rooms, 2 floors] is a real-world dataset collected at
the Oxford Robotics Institute and it includes offices, staircases and a kitchen.
It was collected using a multi-sensor unit consisting of the Sevensense
Alphasense Multi-Camera kit (\figref{fig:system-overview}) integrated with a Hesai
Pandar LiDAR.
\item \emph{Home} [118m, 7 rooms, 2 floors] is a dataset collected from a home
environment, including kitchen, bedrooms, bathroom, living and dining areas, and
a garden. This dataset was recorded and labeled using the same approach as the
\emph{ORI}. 
\end{itemize}

For both \emph{ORI} and \emph{Home}, the LiDAR sensor was used to generate
ground truth. It was not used in LEXIS. Ground truth trajectories were
determined via LiDAR ICP registration against prior maps built with a Leica
BLK360, room labels were hand-labeled using the LiDAR map.

\subsection{Results}


\subsubsection{Room Segmentation and Classification} We define classification
accuracy as the ratio of accurately classified nodes relative to the total
number of nodes in a dataset.  A node is considered accurately classified if the
bounding box that it falls into has the same room label as the node itself. The
reported accuracy is an average of five runs. 

\begin{table}[t]
    \vspace{5pt}
    \centering
    \caption{Mean classification accuracy and standard deviation (over 5 runs). Inference time for
    extracting both image and text encodings for the available \gls{clip} models is also provided.
    Models with inference time of over \SI{40}{\milli\second} were disregarded from
    further analysis.}\label{tab:models}
    \begin{tabular}{@{}llcc@{}}
    \toprule
    \multicolumn{1}{c}{} & \multicolumn{1}{c}{\emph{Home} (\%)} &
    \begin{tabular}[c]{@{}c@{}}\emph{uH2-Apt} (\%)\end{tabular} &
    \begin{tabular}[c]{@{}c@{}}Inference time (ms)\end{tabular} \\ \midrule RN50 &
    73.40\,±\,6.42                  & 53.77\,±\,1.56 & 21.62\,±\,0.04 \\
    \cmidrule(r){1-1} RN101 & 70.66\,±\,2.65                  & 52.49\,±\,3.98 &
    27.28\,±\,0.75 \\ \cmidrule(r){1-1} RN50x4 & 74.58\,±\,5.40 & 55.12\,±\,5.75 &
    28.37\,±\,1.53 \\ \cmidrule(r){1-1} RN50x16 & 75.85\,±\,2.97 &
    \textbf{57.36\,±\,1.27} & 37.14\,±\,0.37 \\ \cmidrule(r){1-1} RN50x64 &
    \multicolumn{1}{c}{-}         & - & 63.57\,±\,0.25 \\ \cmidrule(r){1-1} ViT-B/32
    & 74.96\,±\,4.92                  & 55.51\,±\,2.05 & 20.74\,±\,0.17 \\
    \cmidrule(r){1-1} ViT-B/16 & 77.55\,±\,3.66 & 56.90\,±\,3.51 & 22.36\,±\,0.09 \\
    \cmidrule(r){1-1} ViT-L/14 & \textbf{78.92\,±\,3.01}         &
    \textbf{57.47\,±\,1.81} & 35.61\,±\,0.78 \\ \cmidrule(r){1-1} ViT-L/14@336px &
    \multicolumn{1}{c}{-}         & - & 46.41\,±\,0.25 \\ \bottomrule
    \end{tabular}
    \end{table}

An evaluation of room classification accuracy on a real dataset (\emph{Home})
and a simulated one (\emph{uH2-Apt}) using the available \gls{clip} models is
shown in Table~\ref{tab:models}. RN50x64 and ViT-L/14@336px were excluded due to
their larger size and longer inference times. Refinement parameters $C$ and
$K$ were tuned for each model and dataset, with the best performing models being RN50x16
and ViT-L/14. For further evaluation, we selected RN50x16 for the uHumans2
datasets, as it required less refinement and more effectively preserved
small-scale changes in open-floor plans compared to ViT-L/14. We used ViT-L/14
for all evaluations on the \emph{Home} and \emph{ORI} datasets.
\begin{table}[t]
    \vspace{5pt}
    \centering
    \caption{Mean room classification accuracy and standard deviation over 5
    runs.}\label{tab:hydra_comparison}
    \begin{tabular}{@{}ccccc@{}}
    \toprule
    \%                                                           &
    \begin{tabular}[c]{@{}c@{}}\emph{uH2-Apt}\end{tabular} &
    \begin{tabular}[c]{@{}c@{}}\emph{uH2-Off}\end{tabular} & \emph{ORI}
    & \emph{Home}         \\
    \midrule
    \begin{tabular}[c]{@{}c@{}}Hydra - \\ HRNet\end{tabular}     & 38.0\,±\,21.7
    & 28.4\,±\,6.9 & -            & -            \\
    \begin{tabular}[c]{@{}c@{}}Hydra - \\ OneFormer\end{tabular} & 45\,±\,11.2 &
    27.0\,±\,10.1 & -            & -            \\
    \begin{tabular}[c]{@{}c@{}}LEXIS -\\ Baseline\end{tabular}   &
    51.31\,±\,3.24 & 68.99\,±\,1.07 & 68.09\,±\,1.64 & 61.21\,±\,1.12 \\
    \begin{tabular}[c]{@{}c@{}}LEXIS - \\ Refined\end{tabular}   &
    \textbf{57.36\,±\,1.27} & \textbf{76.03\,±\,1.69} & \textbf{79.22\,±\,4.23}
    & \textbf{78.92\,±\,3.01} \\ \bottomrule
    \end{tabular}
    \end{table} 

Table~\ref{tab:hydra_comparison} presents a comparison of LEXIS using both the
initial segmentation from \gls{clip} (LEXIS - Baseline) and the refined outcome
(LEXIS - Refined) with Hydra~\cite{Hughes2023}. Hydra employs two different 2D semantic segmentation models, using the ADE20k dataset label space~\cite{Zhou2017}. The two segmentation models are HRNet~\cite{Wang2019} and OneFormer~\cite{jain2022oneformer}. We use the same hand-labeled ground-truth bounding boxes as Hydra, but please note that Hydra's evaluation method computes accuracy over clustered room nodes as opposed to the pose nodes used in our method. 

Although the evaluation method differs slightly and it is difficult to precisely compare, our results still show a marked improvement on both Hydra variants. Across the datasets, the refinement procedure improves classification accuracy by an average of 10\%. The key advantage of our open-vocabulary approach is its ability to avoid the constraints of fixed class sets, facilitating effective
generalization to diverse environments and accurate segmentation of open-floor
plans using semantics rather than geometry. For example, in \emph{uH2-Apt}, the
algorithm successfully segmented the \texttt{living room} and \texttt{dining
room} despite the open floor-plan (\figref{fig:refinement}). Similarly, within
the \emph{ORI} dataset, LEXIS divided the \texttt{kitchen} area into
\texttt{kitchen} and \texttt{office} spaces as it contains typical kitchen
equipment as well as whiteboards and tables (\figref{fig:main_fig}).

The variation in the performance of LEXIS on \emph{uH2-Apt} and \emph{uH2-Off}
can be attributed to the ground-truth bounding boxes provided with the dataset,
where stairs are not considered a separate class but instead included as part of
the \texttt{dining room}. Moreover, we hypothesize that the performance
difference between Hydra and LEXIS on the \emph{uH2-Off} dataset can be
attributed to Hydra considering only the object-room graph in training and classification. For instance, in the \emph{uH2-Off} dataset, objects such as chairs and water dispensers are positioned within the corridors. This can be misclassified as \texttt{office} when only considering the objects in the scene, especially if the objects are out-of-distribution from the original training set.

\begin{figure}[t]
    \centering
    \includegraphics[width=\columnwidth]{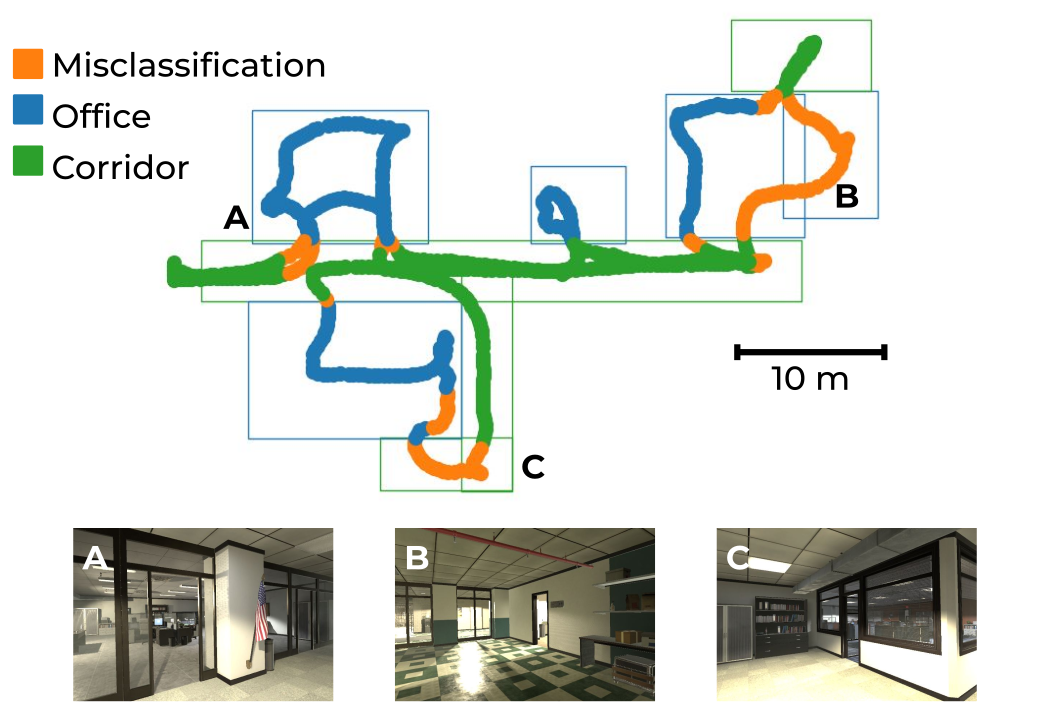}
    \caption{\small{Segmentations produced by LEXIS for the uHumans2 office
    (\emph{uH2-Off}) dataset. Also shown are the ground-truth bounding boxes used in Hydra's evaluation.
    Misclassifications occur during room transitions (example A and B); or areas
    with fewer features (C).}}
    \label{fig:misclassifications}
    \vspace{-10pt}
\end{figure}

Our results on the \emph{uH2-Off} dataset are visualized in
Fig.~\ref{fig:misclassifications}, with orange regions \orangesquare{}
indicating misclassifications. Incorrect classifications are typically clustered
around room edges, e.g., when the camera faces into a room but is actually
located within a corridor (Example A). 

\subsubsection{Semantic Place Recognition} We compared LEXIS' place
recognition method to DBoW~\cite{GalvezLopez2012}, and
NetVLAD~\cite{Arandjelovic2016}. DBoW provides a framework for
feature quantization and indexing of large-scale visual vocabularies. We fed
DBoW with ORB features~\cite{Rublee2011}, as used in the place recognition
systems of ORB-SLAM~\cite{MurArtal2015} and Hydra~\cite{Hughes2022}. NetVLAD is
a neural network architecture pre-trained on Pitts30k~\cite{Torii2013}. 

\begin{figure}[b]
    \centering
    \includegraphics[width=\columnwidth]{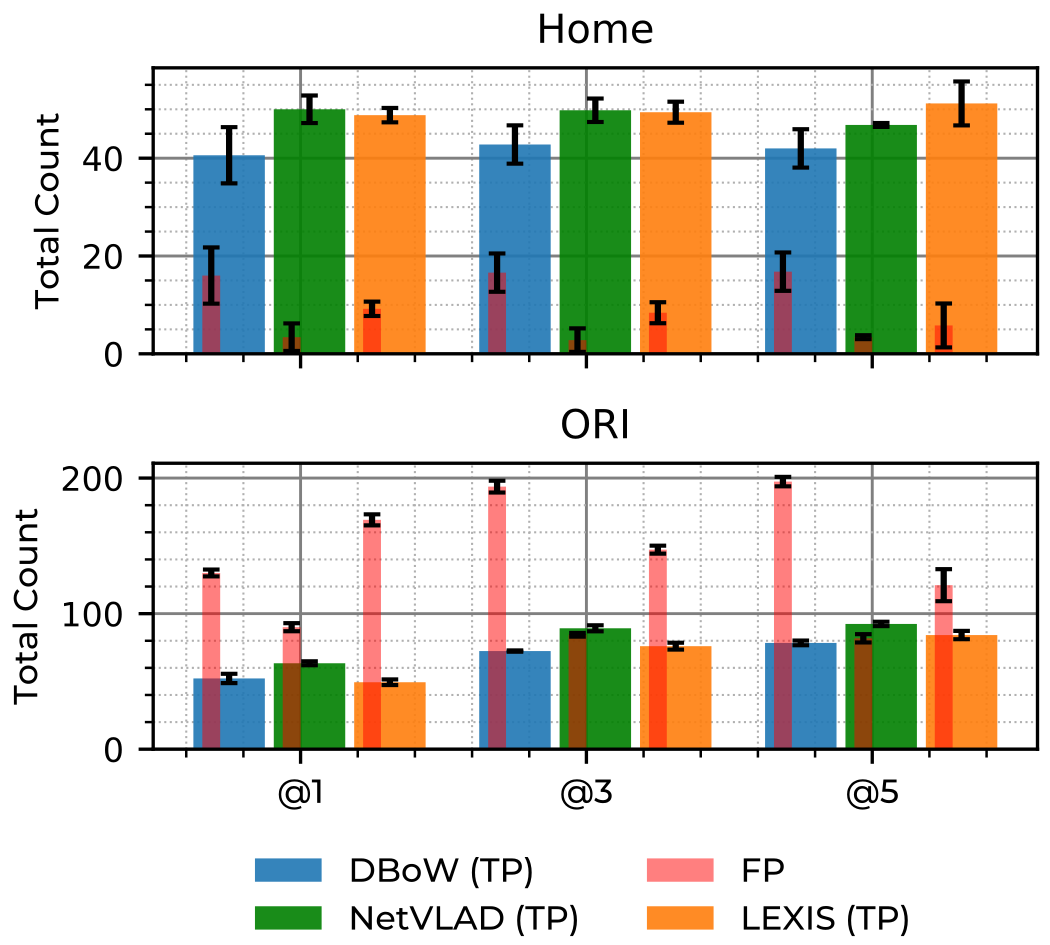}
    \caption{\small{Number of true positives and false positives (red
    \redsquare) using three different VPR methods: DBoW, NetVLAD and LEXIS on
    the \emph{Home} (left) and \emph{ORI} (right) dataset  averaged over 5
    runs.}}\label{fig:vpr_home_plot3}
\end{figure}

We evaluated performance by counting true positives and false positives
(\figref{fig:vpr_home_plot3}). For each query, if $N$ matches were situated
within a distance/angle threshold, we counted it as a true positive; otherwise,
a false positive count was registered. We conducted evaluations at $N$ = 1, 3,
and 5. We used the \emph{Home} and \emph{ORI} datasets, with the true positive
distance threshold set at \SI{1}{\meter} and angular threshold of
\SI{0.5}{\radian}.

As illustrated in \figref{fig:vpr_home_plot3}, our approach achieved more true
positives and less false positives than DBoW across both datasets. The increased
number of true positives can be attributed to the refinement of our search for
loop closures to a relevant room or corridor. 

\begin{figure}[t]
    \vspace{5pt}
    \centering
    \includegraphics[width=\columnwidth]{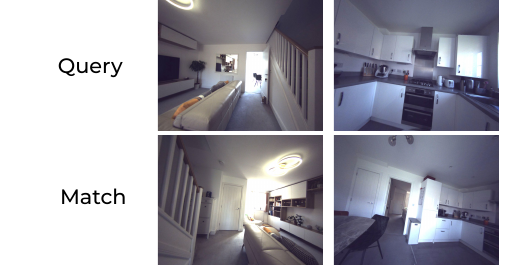}
    \caption{\small{Examples of loop closures provided by CLIP in the \emph{Home}
    dataset. CLIP is able to provide matches from opposing viewpoints (left) and
    with significant viewpoint variations (right) as it relies on semantic
    information.}}\label{fig:clip_matches}
    \vspace{-5pt}
\end{figure}

Interestingly, our method retrieved a similar number of true positives to
NetVLAD despite relying on \gls{clip}, a pre-trained model, with no specific
training for place recognition. We also found that LEXIS produced a slightly
higher number of false positives than NetVLAD. This is primarily due to the
viewpoint variations in the suggested matches produced by our method, as
demonstrated in \figref{fig:clip_matches}. Notably, due to \gls{clip} relying
solely on semantic information, opposing viewpoints are presented as potential
matches.  

The high number of false positives across all methods in the \emph{ORI} dataset
could be attributed to there being many visually similar offices in the dataset. However, it
is worth noting, particularly with robust graph optimization techniques and PnP
verification, prioritizing the accurate identification of enough valid loop
closures (true positives) is more important than avoiding incorrect loop
closures (false positives)~\cite{schubert2023visual}.

\subsubsection{Full System Evaluation} We conducted a comparison using LEXIS as
a complete \gls{slam} system, benchmarked against two state-of-the-art
alternatives: ORB-SLAM3~\cite{ORBSLAM3_TRO} and VINS-Fusion~\cite{qin2017vins}.
In our experiments, we used the stereo-inertial configurations with loop
closures enabled and assessed performance using the \gls{ate}. The results are
summarized in Table~\ref{tab:ate}.

\begin{table}[h]
    \centering
    \caption{Comparison of \gls{ate} in the \emph{ORI} and \emph{Home}
    datasets.}\label{tab:ate}
    \begin{tabular}{@{}lll@{}}
    \toprule
    ATE (m)     & \emph{ORI} & \emph{Home} \\ \midrule ORB-SLAM3   & 0.22 & 0.10
    \\
    VINS-Fusion & 0.10 & 0.08 \\
    LEXIS       & 0.16 & 0.10 \\ \bottomrule
    \end{tabular}
    \end{table}

Despite the streamlined and minimal design of LEXIS --- combining the Multi-Camera VILENS VIO
system~\cite{ZhangWCF22}, with classical pose graph optimization and our
\gls{clip}-based semantic place recognition module, it still achieves comparable
performance to that of ORB-SLAM3 and VINS-Fusion. The incorporation of the
Multi-Camera system, which analyzes images from two front-facing and two
lateral-facing cameras, provides benefits as the system can avoid tracking
issues in confined indoor environments.

\subsubsection{Planning Application} 
\begin{figure}[h]
    \centering
    \includegraphics[width=\columnwidth, trim=0cm 0.5cm 0cm 1cm, clip]{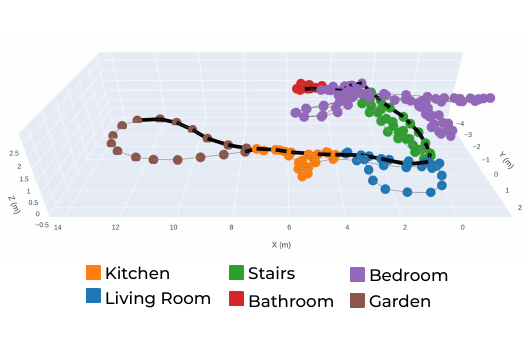}
    \caption{\small{Segmentation of the \emph{Home} dataset with a topological
    plan, shown in black $\blacksquare$, from the \texttt{bathroom} to the
    \texttt{garden}.}}\label{fig:planning}
\end{figure}

Finally, we demonstrated that the 
representation produced by LEXIS can be used for mission planning in a
real-world environment encompassing multiple floors and rooms. From the pose
graph, we constructed an adjacency matrix that establishes connections between
consecutive nodes and nodes within the same cluster. We then computed the
shortest path between initial and goal room labels using Dijkstra's
algorithm~\cite{dijkstra1959note}. An example path on the \emph{Home} dataset is
illustrated in \figref{fig:planning}.

\section{Conclusion}\label{sec:conclusion}

This work presents LEXIS, a real-time semantic visual \gls{slam} system enhanced
by open-vocabulary language models. Our system constructs a topological model of
indoor environments that is enriched with embedded semantic understanding. This
allows us to properly segment rooms and spaces across diverse contexts.
Leveraging this representation, we demonstrated room-aware place recognition
which achieves performance equivalent with established place recognition methods
such as NetVLAD and DBoW. We evaluated our \gls{slam} system in home and office
environments and achieved comparable \gls{ate} to established systems (ORB-SLAM3
and VINS-Fusion). Finally, we demonstrated an example of how our representation
can be used for other robotics tasks such as room-to-room planning. This work
showcases how open-vocabulary models can enable autonomous systems to interact
naturally with their environment. Future work will focus on enhancing room
classification by integrating LEXIS with dense reconstruction techniques and
considering uncertainty in the estimation for long term use of the system. We
also intend to investigate per-pixel adaptations of the \gls{clip} model.

\section*{Acknowledgments}
This work is supported in part by a Royal Society University Research Fellowship
(Fallon, Kassab), and the ANID/BECAS CHILE/2019-72200291 (Mattamala). We thank
Nathan Hughes for providing the ground truth for the uHumans2 dataset. For the
purpose of Open Access, the author has applied a CC BY public copyright licence
to any Author Accepted Manuscript (AAM) version arising from this submission.

\balance
\bibliographystyle{IEEEtran}
\bibliography{references}

\begin{thebibliography}{10}
\providecommand{\url}[1]{#1}
\csname url@rmstyle\endcsname
\providecommand{\newblock}{\relax}
\providecommand{\bibinfo}[2]{#2}
\providecommand\BIBentrySTDinterwordspacing{\spaceskip=0pt\relax}
\providecommand\BIBentryALTinterwordstretchfactor{4}
\providecommand\BIBentryALTinterwordspacing{\spaceskip=\fontdimen2\font plus
\BIBentryALTinterwordstretchfactor\fontdimen3\font minus \fontdimen4\font\relax}
\providecommand\BIBforeignlanguage[2]{{%
\expandafter\ifx\csname l@#1\endcsname\relax
\typeout{** WARNING: IEEEtran.bst: No hyphenation pattern has been}%
\typeout{** loaded for the language `#1'. Using the pattern for}%
\typeout{** the default language instead.}%
\else
\language=\csname l@#1\endcsname
\fi
#2}}

\bibitem{Cadena2016}
C.~Cadena, L.~Carlone, H.~Carrillo, Y.~Latif, D.~Scaramuzza, J.~Neira, I.~Reid, and J.~J. Leonard, ``{Past, Present, and Future of Simultaneous Localization and Mapping: Toward the Robust-Perception Age},'' \emph{{IEEE} Trans. Robotics}, vol.~32, no.~6, pp. 1309--1332, 2016.

\bibitem{Davison2018}
A.~J. Davison, ``{FutureMapping: The Computational Structure of Spatial AI Systems},'' \emph{CoRR}, vol. abs/1803.11288, 2018.

\bibitem{Armeni2019}
I.~Armeni, Z.-Y. He, J.~Gwak, A.~R. Zamir, M.~Fischer, J.~Malik, and S.~Savarese, ``{3d scene graph: A structure for unified semantics, 3d space, and camera},'' in \emph{{IEEE} Int. Conf. Computer Vision and Pattern Recognition}, 2019, pp. 5664--5673.

\bibitem{Kim2019}
U.-H. Kim, J.-M. Park, T.-j. Song, and J.-H. Kim, ``{3-D Scene Graph: A Sparse and Semantic Representation of Physical Environments for Intelligent Agents},'' \emph{IEEE Trans. Cybern.}, vol.~50, no.~12, pp. 4921--4933, 2020.

\bibitem{Hughes2022}
N.~Hughes, Y.~Chang, and L.~Carlone, ``{Hydra: A Real-time Spatial Perception System for 3D Scene Graph Construction and Optimization},'' in \emph{Robotics: Science and Systems (RSS)}, 2022.

\bibitem{Bavle2023}
H.~Bavle, J.~L. Sanchez-Lopez, M.~Shaheer, J.~Civera, and H.~Voos, ``{S-Graphs+: Real-Time Localization and Mapping Leveraging Hierarchical Representations},'' \emph{{IEEE} Robot. Autom. Lett. (RA-L)}, vol.~8, no.~8, pp. 4927--4934, 2023.

\bibitem{Radford2021}
A.~Radford, J.~W. Kim, C.~Hallacy, A.~Ramesh, G.~Goh, S.~Agarwal, G.~Sastry, A.~Askell, P.~Mishkin, J.~Clark, G.~Krueger, and I.~Sutskever, ``{Learning Transferable Visual Models From Natural Language Supervision},'' \emph{CoRR}, vol. abs/2103.00020, 2021.

\bibitem{Gu2022}
X.~Gu, T.~Lin, W.~Kuo, and Y.~Cui, ``{Open-vocabulary Object Detection via Vision and Language Knowledge Distillation},'' in \emph{Intl. Conf. on Learning Representations (ICLR)}, 2022.

\bibitem{Peng2023}
S.~Peng, K.~Genova, C.~M. Jiang, A.~Tagliasacchi, M.~Pollefeys, and T.~Funkhouser, ``{OpenScene: 3D Scene Understanding with Open Vocabularies},'' in \emph{{IEEE} Int. Conf. Computer Vision and Pattern Recognition}, 2023.

\bibitem{Jatavallabhula2023}
K.~M. Jatavallabhula, A.~Kuwajerwala, Q.~Gu, M.~Omama, T.~Chen, S.~Li, G.~Iyer, S.~Saryazdi, N.~Keetha, A.~Tewari, J.~B. Tenenbaum, C.~M. de~Melo, M.~Krishna, L.~Paull, F.~Shkurti, and A.~Torralba, ``{ConceptFusion: Open-set Multimodal 3D Mapping},'' in \emph{Robotics: Science and Systems (RSS)}, 2023.

\bibitem{Chen2023}
B.~Chen, F.~Xia, B.~Ichter, K.~Rao, K.~Gopalakrishnan, M.~S. Ryoo, A.~Stone, and D.~Kappler, ``{Open-vocabulary Queryable Scene Representations for Real World Planning},'' in \emph{IEEE Int. Conf. Robot. Autom. (ICRA)}, 2023, pp. 11\,509--11\,522.

\bibitem{Huang2023}
C.~Huang, O.~Mees, A.~Zeng, and W.~Burgard, ``{Visual Language Maps for Robot Navigation},'' in \emph{IEEE Int. Conf. Robot. Autom. (ICRA)}, 2023.

\bibitem{Rottmann2005}
A.~Rottmann, O.~Mozos, C.~Stachniss, and W.~Burgard, ``{Semantic Place Classification of Indoor Environments with Mobile Robots Using Boosting.}'' in \emph{Proceedings of the National Conference on Artificial Intelligence}, vol.~3, 01 2005, pp. 1306--1311.

\bibitem{Goeddel2016}
R.~Goeddel and E.~Olson, ``{Learning semantic place labels from occupancy grids using CNNs},'' in \emph{IEEE/RSJ Intl. Conf. on Intelligent Robots and Systems (IROS)}, 2016, pp. 3999--4004.

\bibitem{sunderhauf2015place}
N.~Sünderhauf, F.~Dayoub, S.~McMahon, B.~Talbot, R.~Schulz, P.~Corke, G.~Wyeth, B.~Upcroft, and M.~Milford, ``{Place Categorization and Semantic Mapping on a Mobile Robot},'' \emph{CoRR}, vol. abs/1507.02428, 2015.

\bibitem{Wang2019}
J.~Wang, K.~Sun, T.~Cheng, B.~Jiang, C.~Deng, Y.~Zhao, D.~Liu, Y.~Mu, M.~Tan, X.~Wang, W.~Liu, and B.~Xiao, ``{Deep High-Resolution Representation Learning for Visual Recognition},'' \emph{CoRR}, vol. abs/1908.07919, 2019.

\bibitem{Hughes2023}
N.~Hughes, Y.~Chang, S.~Hu, R.~Talak, R.~Abdulhai, J.~Strader, and L.~Carlone, ``{Foundations of Spatial Perception for Robotics: Hierarchical Representations and Real-time Systems},'' \emph{CoRR}, vol. abs/2305.07154, 2023.

\bibitem{kerr2023lerf}
J.~Kerr, C.~M. Kim, K.~Goldberg, A.~Kanazawa, and M.~Tancik, ``{LERF: Language Embedded Radiance Fields},'' \emph{CoRR}, vol. abs/2303.09553, 2023.

\bibitem{shafiullah2022clipfields}
N.~M.~M. Shafiullah, C.~Paxton, L.~Pinto, S.~Chintala, and A.~Szlam, ``{CLIP-Fields: Weakly Supervised Semantic Fields for Robotic Memory},'' in \emph{Robotics: Science and Systems (RSS)}, 2023.

\bibitem{Jatavallabhula2020}
K.~M. Jatavallabhula, G.~Iyer, and L.~Paull, ``{Grad SLAM: Dense SLAM meets Automatic Differentiation},'' in \emph{IEEE Int. Conf. Robot. Autom. (ICRA)}, 2020, pp. 2130--2137.

\bibitem{ha2022semantic}
H.~Ha and S.~Song, ``{Semantic Abstraction: Open-World 3D Scene Understanding from 2D Vision-Language Models},'' \emph{CoRR}, vol. abs/2207.11514, 2022.

\bibitem{ZhangWCF22}
L.~Zhang, D.~Wisth, M.~Camurri, and M.~F. Fallon, ``{Balancing the Budget: Feature Selection and Tracking for Multi-Camera Visual-Inertial Odometry},'' \emph{{IEEE} Robot. Autom. Lett. (RA-L)}, vol.~7, no.~2, pp. 1182--1189, 2022.

\bibitem{Alcantarilla2012}
P.~F. Alcantarilla, A.~Bartoli, and A.~J. Davison, ``{KAZE Features},'' in \emph{Eur. Conf. on Computer Vision (ECCV)}, vol. 7577, 2012, pp. 214--227.

\bibitem{Zhu2022}
X.~Zhu and Z.~Ghahramani, ``{Learning from Labeled and Unlabeled Data with Label Propagation},'' Carnegie Mellon University, Tech. Rep., 2002.

\bibitem{Fischler1981}
M.~A. Fischler and R.~C. Bolles, ``{Random Sample Consensus: A Paradigm for Model Fitting with Applications to Image Analysis and Automated Cartography},'' \emph{Commun. ACM}, vol.~24, no.~6, p. 381–395, 1981.

\bibitem{Agarwal2013}
P.~Agarwal, G.~D. Tipaldi, L.~Spinello, C.~Stachniss, and W.~Burgard, ``{Robust map optimization using dynamic covariance scaling},'' in \emph{IEEE Int. Conf. Robot. Autom. (ICRA)}, 2013, pp. 62--69.

\bibitem{Rosinol2021}
A.~Rosinol, A.~Violette, M.~Abate, N.~Hughes, Y.~Chang, J.~Shi, A.~Gupta, and L.~Carlone, ``{Kimera: from SLAM to Spatial Perception with 3D Dynamic Scene Graphs},'' \emph{Intl. J. of Robot. Res.}, 2021.

\bibitem{Zhou2017}
B.~Zhou, H.~Zhao, X.~Puig, S.~Fidler, A.~Barriuso, and A.~Torralba, ``{Scene Parsing through ADE20K Dataset},'' in \emph{{IEEE} Int. Conf. Computer Vision and Pattern Recognition}, 2017, pp. 5122--5130.

\bibitem{jain2022oneformer}
J.~Jain, J.~Li, M.~Chiu, A.~Hassani, N.~Orlov, and H.~Shi, ``{OneFormer: One Transformer to Rule Universal Image Segmentation},'' in \emph{{IEEE} Int. Conf. Computer Vision and Pattern Recognition}, 2023, pp. 2989--2998.

\bibitem{GalvezLopez2012}
D.~Galvez-López and J.~D. Tardos, ``{Bags of Binary Words for Fast Place Recognition in Image Sequences},'' \emph{{IEEE} Trans. Robotics}, vol.~28, no.~5, pp. 1188--1197, 2012.

\bibitem{Arandjelovic2016}
R.~Arandjelovic, P.~Gronat, A.~Torii, T.~Pajdla, and J.~Sivic, ``{NetVLAD}: {CNN} architecture for weakly supervised place recognition,'' in \emph{{IEEE} Int. Conf. Computer Vision and Pattern Recognition}, 2016, pp. 5297--5307.

\bibitem{Rublee2011}
E.~Rublee, V.~Rabaud, K.~Konolige, and G.~R. Bradski, ``{ORB:} an efficient alternative to {SIFT} or {SURF},'' in \emph{Intl. Conf. on Computer Vision (ICCV)}, D.~N. Metaxas, L.~Quan, A.~Sanfeliu, and L.~V. Gool, Eds., 2011, pp. 2564--2571.

\bibitem{MurArtal2015}
R.~Mur-Artal, J.~M.~M. Montiel, and J.~D. Tardós, ``{ORB-SLAM: A Versatile and Accurate Monocular SLAM System},'' \emph{{IEEE} Trans. Robotics}, vol.~31, no.~5, pp. 1147--1163, 2015.

\bibitem{Torii2013}
A.~Torii, J.~Sivic, T.~Pajdla, and M.~Okutomi, ``{Visual Place Recognition with Repetitive Structures},'' in \emph{{IEEE} Int. Conf. Computer Vision and Pattern Recognition}, 2013, pp. 883--890.

\bibitem{schubert2023visual}
S.~Schubert, P.~Neubert, S.~Garg, M.~Milford, and T.~Fischer, ``{Visual Place Recognition: A Tutorial},'' \emph{CoRR}, vol. abs/2303.03281, 2023.

\bibitem{ORBSLAM3_TRO}
C.~Campos, R.~Elvira, J.~J. G\'omez, J.~M.~M. Montiel, and J.~D. Tard\'os, ``{ORB-SLAM3}: An accurate open-source library for visual, visual-inertial and multi-map {SLAM},'' \emph{{IEEE} Trans. Robotics}, vol.~37, no.~6, pp. 1874--1890, 2021.

\bibitem{qin2017vins}
T.~Qin, P.~Li, and S.~Shen, ``Vins-mono: A robust and versatile monocular visual-inertial state estimator,'' \emph{{IEEE} Trans. Robotics}, vol.~34, no.~4, pp. 1004--1020, 2018.

\bibitem{dijkstra1959note}
E.~W. Dijkstra, ``{A note on two problems in connexion with graphs},'' \emph{Numerische mathematik}, vol.~1, no.~1, pp. 269--271, 1959.

\end{thebibliography}


\end{document}